\documentclass[11pt]{article}

\usepackage[final]{acl}

\usepackage{times}
\usepackage{latexsym}

\usepackage[T1]{fontenc}

\usepackage[utf8]{inputenc}

\usepackage{microtype}

\usepackage{inconsolata}

\usepackage{graphicx}
\usepackage{booktabs}
\usepackage{float}
\usepackage{amsmath}
\usepackage{amssymb}


\title{Usage-Modulated Sentiment Representations in Large Language Models}

\author{
  \textbf{Hongfei Du},
  \textbf{Jiacheng Shi},
  \textbf{Yanfu Zhang},
  \textbf{Gang Zhou},
  \textbf{Ye Gao}
  \\
  {\normalfont Department of Computer Science, William \& Mary, Williamsburg, VA, USA}
  \\
  {\normalfont\texttt{\{hdu02, jshi12, yzhang105, gzhou, ygao18\}@wm.edu}}
}

\begin{document}
\maketitle
\begin{abstract}
Prior work suggests that sentiment can often be captured by approximately
linear directions in LLM activation spaces, but a single direction may not
fully capture sentiment representations. In natural communication,
sentiment is shaped not only by polarity but also by usage factors, such as
tone and audience adaptation. We test whether these factors systematically
modulate sentiment representations beyond a shared sentiment direction. We
construct a controlled paired dataset that holds event content fixed while
varying sentiment polarity and usage factors, and analyze Llama, Mistral, and
Gemma. We identify a shared sentiment direction, remove it, and test the
residual structure through erasure and generation-time tone steering. Across
models, the shared direction is robust (median cosine 0.953--0.975), yet
removing it leaves 0.833--0.909 of the original positive--negative
representation-difference norm. The residuals contain compact, reproducible
usage-conditioned structure. Targeted erasure weakens held-out usage metrics
more than random and label-shuffled controls. On Llama, outputs steered along
residualized tone components are preferred in 92.8\% of blind target-tone
comparisons while preserving the requested sentiment polarity in 98.7\% of
evaluated outputs.
\end{abstract}

\section{Introduction}

Large language models (LLMs) have rapidly advanced capabilities in language
understanding and generation~\citep{radford2019language,brown2020language}.
They are increasingly applied in domains where sentiment is central, such as
mental health support and psychotherapy assistance~\citep{
yang2023towards,gabriel2024can,hu2025beyond}. These uses motivate studying
sentiment not only as output polarity, but also as an internal representation.

A common approach to representation analysis is grounded in the linear
representation hypothesis: high-level properties can often be captured as
directions in activation
space~\citep{mikolov2013linguistic,park2023linear,jiang2024origins}. This
approach has been used to study abstract concepts such as truth, refusal, and
persona~\citep{marks2023geometry,arditi2024refusal,chen2025persona}. Prior work
has identified a single sentiment direction and established its causal role
through activation interventions~\citep{hollinsworth-etal-2024-language}, while
layerwise probing has shown that sentiment information can be decoded by linear
classifiers across layers and model scales within the LLaMA
family~\citep{palma2025llamas}. However, a single sentiment direction may not
fully capture sentiment representations: single
concept vectors can vary with data and probe training, motivating subspace
views of concepts~\citep{zhao2025beyond}. In natural communication, sentiment
is shaped not only by polarity but also by usage factors, such as tone and
audience adaptation~\citep{clark1982audience,bell1984language}. We therefore
explore whether usage modulates sentiment representations beyond a shared
sentiment direction, rather than being noise.

To answer this question, we construct a controlled paired dataset that holds
event content fixed while varying sentiment polarity and usage factors.
Natural text makes this hard: positive and negative examples often differ not
only in sentiment, but also in topic, wording, audience, and pragmatic context.
We instead realize each controlled event as matched negative and positive texts
under the same usage value. We refer to the resulting
within-event positive--negative representation difference as a
\textit{sentiment delta}.

\begin{figure*}[t]
    \centering
    \includegraphics[width=\textwidth]{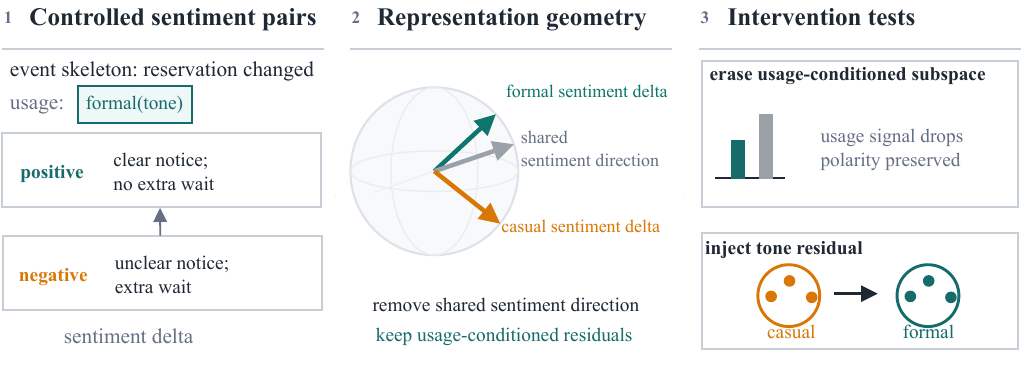}
    \caption{Controlled analysis pipeline. For each event skeleton and usage
    value, we encode matched positive and negative realizations to form a
    sentiment delta. We separate a shared sentiment direction from
    usage-conditioned residual structure, then test this residual structure through
    targeted erasure and generation-time tone steering. The example illustrates tone
    values; audience values are treated analogously.}
    \label{fig:method-overview}
\end{figure*}

Figure~\ref{fig:method-overview} summarizes our controlled analysis pipeline.
Within this controlled setting, our analysis first separates a shared sentiment
direction from usage-conditioned residual structure, then summarizes the residual
structure with compact subspaces and tests it with interventions.
\textbf{Shared sentiment direction:} we first identify a direction shared by
sentiment deltas and test whether it generalizes to unseen event skeletons.
\textbf{Usage-conditioned residual structure:} we remove this shared direction
and test whether the residual variation remains organized by communicative
usage after shuffled-label baselines; we then summarize this structure with
compact tone and audience subspaces and test whether these subspaces reproduce
when estimated independently from disjoint training and test skeleton sets.
\textbf{Intervention tests:} we erase the residual usage subspaces and test
whether injecting residualized usage components can steer Llama generations
toward a target tone or audience while preserving sentiment polarity.

The results show systematic usage modulation beyond the shared sentiment
direction. Across all three models, sentiment deltas contain a robust shared
sentiment direction
(median cosine 0.953--0.975), but this direction does not fully explain the
representation change. After removing it, much of the original norm remains
(0.833--0.909), and the residuals retain compact, reproducible
usage-conditioned structure. Targeted erasure selectively weakens this
structure, and generation experiments show that residualized tone components
can steer Llama output tone while preserving sentiment polarity in 98.7\% of
evaluated prompts. Independent judging corroborates the tone-steering result,
while held-out human-written and audience evaluations provide complementary
evidence of transfer.

Our contributions are as follows:
\begin{itemize}
    \item We build a controlled sentiment dataset that instantiates the same
    event skeleton under both sentiment polarities and different tone and
    audience values,
    allowing us to compare sentiment representations while holding event content
    fixed.
    \item We show that sentiment representations are not captured by a single
    direction. Across Llama, Mistral, and Gemma, a shared sentiment direction is
    robust, but removing it leaves substantial, stable, usage-conditioned
    residual structure.
    \item We test whether this residual structure matters through interventions.
Residual erasure reduces tone and audience structure more than control
erasures, and residualized tone components steer the tone of Llama outputs at
generation time while preserving sentiment polarity.
\end{itemize}

\begin{table*}[!t]
\centering
\small
\setlength{\tabcolsep}{2.5pt}
\renewcommand{\arraystretch}{1.0}
\begin{tabular}{@{}llp{0.37\textwidth}p{0.37\textwidth}@{}}
\toprule
Family & Value & Positive realization & Negative realization \\
\midrule
Tone & formal &
The restaurant efficiently notified patrons of the reservation time adjustment
on Monday, ensuring the process was seamless and without delay. &
The restaurant's notification regarding the reservation time adjustment on
Monday was inadequately communicated, resulting in unnecessary delays for the
patrons. \\
Tone & casual &
The restaurant let us know about the time change on Monday, and it was super
smooth without any extra waiting. &
We got a confusing message about the time change on Monday, and it ended up
making us wait longer. \\
\addlinespace[1pt]
Audience & specialist-facing &
The restaurant adjusted the reservation time on Monday, ensuring the change was
communicated effectively and managed seamlessly, eliminating any additional
waiting. &
On Monday, the restaurant altered the reservation time, but the unclear
communication led to additional waiting for patrons. \\
Audience & child-friendly &
The restaurant told everyone about the new time for tables on Monday, and nobody
had to wait longer. &
The restaurant changed the time for tables on Monday, but they didn't tell
everyone well, so people had to wait longer. \\
\bottomrule
\end{tabular}
\caption{Representative controlled realizations of one reservation-change
event. Rows hold event and usage value fixed while varying outcome polarity
across tone and audience adaptation.}
\label{tab:representative-examples}
\end{table*}

\section{Related Work}

\paragraph{Concept Vectors.}
A common approach to representation analysis is to identify concept vectors:
directions in activation space that track meaningful model-internal properties.
The linear representation hypothesis motivates this view for
LLMs~\citep{park2023linear,jiang2024origins}, and empirical work has found
direction-like structure for truth, refusal, persona, and sentiment
polarity~\citep{marks2023geometry,arditi2024refusal,chen2025persona,
hollinsworth-etal-2024-language,palma2025llamas}. At the same time, single-vector descriptions
can be sensitive to the examples or probes used to construct them, motivating
subspace or distributional views of concepts~\citep{pahde2022navigating,
zhao2025beyond}. Recent work maps broader affective structure in LLMs, including
emotion and valence--arousal representations~\citep{choi2026latent,
sun2026valence}. Our question is more specific: does sentiment structure beyond
a shared sentiment direction track communicative usage in LLMs?

\paragraph{Usage-Conditioned Sentiment Realization}
Sentiment is expressed through more than polarity: the same positive or negative
event can be phrased differently depending on tone and audience. Audience-design
work studies how speakers adapt wording, style, reference, and detail to
addressees~\citep{clark1982audience,bell1984language}, and NLP work treats
style, politeness, and audience as controllable aspects of text
realization~\citep{ficler2017controlling,niu2018polite,
moorjani2022audience,shu2020controllable}. These works motivate tone and
audience as meaningful usage factors, but they do not test whether such factors
structure LLM representations of sentiment. Controlled-comparison datasets,
including counterfactually augmented data and contrast sets, provide a standard
way to vary one factor while holding the surrounding context
fixed~\citep{kaushik2020learning,gardner2020evaluating}. We use this logic at
the representation level: after removing a shared sentiment direction, we ask
whether the remaining sentiment structure still tracks usage.

\paragraph{Activation Steering.}
Activation steering changes model behavior by adding or modifying internal
representation directions during generation~\citep{zou2023representation,
li2023inference,turner2023steering,liu2023incontext,rimsky2024steering,
vanderweij2024extending}. These directions are often derived from contrastive
examples or learned representation differences. Prior work has used
activation-space interventions to improve
truthfulness~\citep{li2023inference}, modulate refusal
behavior~\citep{arditi2024refusal}, control personality
shifts~\citep{chen2025persona}, shift sentiment~\citep{turner2023steering},
and control affective properties~\citep{sun2026valence}. Most steering work
uses a learned direction as a direct control handle for a target attribute. In
contrast, we use steering as a diagnostic: we ask whether usage-conditioned
residual structure can be used to change output tone while preserving the
requested sentiment polarity.

\section{Dataset Construction}
\label{sec:dataset}

We construct the dataset to measure sentiment change while holding event content
and communicative usage fixed. Each event skeleton specifies a neutral event and
parallel positive and negative outcome frames. For every usage value, we realize
the skeleton twice, once with each outcome frame, yielding a controlled
positive--negative pair.

We consider two usage factors: tone and audience adaptation. In the dataset
and analysis, we refer to these as usage families, each containing four usage
values. Tone controls how the same content is worded, with values
\emph{formal}, \emph{casual},
\emph{enthusiastic}, and \emph{restrained}. Audience adaptation controls the
assumed reader, with values \emph{specialist-facing},
\emph{general-public-facing}, \emph{child-friendly}, and \emph{peer-facing}.
These values span common tone and audience contrasts without intending to
exhaust communicative usage.
Table~\ref{tab:representative-examples} illustrates this paired construction
for one event skeleton under representative tone and audience values.

\begin{table}[t]
\centering
\small
\setlength{\tabcolsep}{3pt}
\begin{tabular}{@{}llrrrr@{}}
\toprule
Judge & Usage & $N$ & Assign. & Event & Sent. \\
\midrule
Human & Tone & 40 & 100.0 & 100.0 & 100.0 \\
Human & Audience & 40 & 100.0 & 100.0 & 100.0 \\
\midrule
GPT-4o mini & Tone & 40 & 95.0 & 100.0 & 97.5 \\
GPT-4o mini & Audience & 40 & 92.5 & 100.0 & 92.5 \\
\bottomrule
\end{tabular}
\caption{Blind validation of controlled usage contrasts. $N$ is the number of
items; Assign., Event, and Sent. report correct usage assignment, event
preservation, and sentiment preservation (\%).}
\label{tab:manual-validation}
\end{table}

We use GPT-4o~\citep{openai2024gpt4o} to generate 300 candidate event skeletons, realize them under the requested
usage values, and then apply filtering and completeness checks. We retain the
221 skeletons with complete balanced realizations for both usage families.
Because each retained skeleton is realized under eight usage values and two
sentiment polarities, the final dataset contains
$221 \times 8 \times 2 = 3{,}536$ texts.
Appendix~\ref{app:dataset-details} gives the full usage definitions, generation
protocol, and filtering details.

We validate the retained dataset with automatic checks for malformed rows,
duplicate targets, label leakage, and length or content violations, followed by
blind assignment checks over sampled retained pairs. In each blind check, a
human annotator and an LLM judge assign two same-skeleton texts to a usage contrast
and verify event and sentiment preservation. The validation subsets are
balanced across usage contrasts, sentiment polarities, and the two possible
A/B assignment orders. Judges see the base event, sentiment frame, contrast
definitions, and two unlabeled realizations, but not the hidden usage labels.
They may select either assignment or mark the contrast as tied or unclear, and
they assess event and sentiment preservation separately from usage assignment.
Wording, formality, intensity, or abstraction changes do not count as
preservation failures unless the central event or outcome is changed,
contradicted, or omitted.

Table~\ref{tab:manual-validation} reports blind spot checks on the retained
controlled pairs. Human checks cover tone and audience subsets, and the LLM
checks use GPT-4o mini~\citep{openai2024gpt4omini} with the same assignment
format for additional coverage. Across these checks, intended usage assignment
is recovered in 92.5--100.0\% of items, with high event and sentiment
preservation. A single human annotator provides one judgment for each of the 40
tone items and 40 audience items. We therefore treat these rows as quality-control
audits rather than agreement across multiple annotators, and do not report
inter-annotator agreement. Together with the automatic checks over all 3,536
texts, these blind spot checks support using the retained pairs as a controlled analysis set.
Because GPT-4o mini is used only as secondary audit evidence,
Appendix~\ref{app:judge-disagreements} reports representative LLM-judge
disagreements.

\section{Shared Sentiment Direction and Usage-Conditioned Residual Structure}
\label{sec:shared-residual-geometry}

This section tests whether sentiment representations decompose into a shared
sentiment direction plus usage-conditioned residual structure. We first identify
a sentiment direction shared across tone and audience conditions, then remove
this direction and ask whether the remaining residuals still organize by usage
rather than collapse to noise. We evaluate three LLMs, Llama~\citep{grattafiori2024llama},
Mistral~\citep{jiang2023mistral7b}, and Gemma~\citep{team2024gemma}, and report
results for both mean-pooled and last-token representations.
See Appendix~\ref{app:model-details} for model details.

\begin{table*}[!t]
    \centering
    \small
    \begin{tabular}{@{}llrrrrrr@{}}
    \toprule
    Model & Pooling & Tone align & Aud. align & Proj. share & Resid. ratio & Tone dec. margin & Aud. dec. margin \\
    \midrule
    Llama & mean & 0.962 & 0.969 & 0.188 & 0.899 & 0.170 & 0.049 \\
    Mistral & mean & 0.965 & 0.966 & 0.180 & 0.904 & 0.220 & 0.201 \\
    Gemma & mean & 0.968 & 0.963 & 0.172 & 0.909 & 0.114 & 0.144 \\
    Llama & last & 0.957 & 0.966 & 0.244 & 0.868 & 0.326 & 0.326 \\
    Mistral & last & 0.953 & 0.960 & 0.192 & 0.898 & 0.352 & 0.284 \\
    Gemma & last & 0.975 & 0.975 & 0.301 & 0.833 & 0.356 & 0.280 \\
    \bottomrule
    \end{tabular}
    \caption{Shared sentiment alignment and residual usage structure, summarized
    across layers. Tone align and Aud. align are median cosine alignments between
    the held-out tone or audience mean sentiment delta and the training-set
    sentiment direction. Proj. share is the median sample-level share of squared
    norm captured by that direction over individual test-split sentiment deltas.
    Resid. ratio is the median sample-level remaining norm ratio after projection.
    Tone dec. margin and Aud. dec. margin are nearest-centroid usage-decoding
    margins above the 95th percentile of within-skeleton label shuffles; positive
    values indicate residual usage structure above the shuffle baseline.}
    \label{tab:shared-residual-summary}
\end{table*}

\subsection{Shared Sentiment Direction}
\label{sec:sentiment-deltas}

Let
$\mathcal{S}$ denote the event skeletons in the final dataset, and let
$\mathcal{U}$ denote the eight usage values described in
Section~\ref{sec:dataset}: four tone settings and four audience settings. For
all representation analyses, we split $\mathcal{S}$ by skeleton into
$\mathcal{S}_{\mathrm{train}}$, $\mathcal{S}_{\mathrm{dev}}$, and
$\mathcal{S}_{\mathrm{test}}$ with 155, 33, and 33 skeletons, respectively.
The development split is used to select the steering strength for the
generation experiment; the results in this section are reported on the test
skeletons. Splitting by skeleton prevents
different realizations of the same event from appearing in both training and
test analyses.
For each input text $x$, we run the model forward and extract the hidden states
at layer $\ell$. We convert these token-level hidden states into a single vector
$h^\ell(x)$ in two ways: mean pooling over text tokens and taking the last-token
state. All analyses below are run separately for each model, layer, and pooling
choice.

For skeleton $s$ and usage value $u$, we define the sentiment delta and shared
sentiment direction as
\begin{equation}
    \begin{aligned}
    \Delta^\ell_{s,u} &= h^\ell(x^+_{s,u})-h^\ell(x^-_{s,u}), \\
    B^\ell &= \frac{\sum_{s,u}\Delta^\ell_{s,u}}
    {\left\|\sum_{s,u}\Delta^\ell_{s,u}\right\|}.
    \end{aligned}
\end{equation}
Here $x^+_{s,u}$ and $x^-_{s,u}$ are the positive and negative realizations of
the same event skeleton under the same usage value. Thus $B^\ell$ captures
polarity-aligned change common across usage values rather than a direction tied
to a particular tone or audience, with the sums over training skeletons and all
usage values.
Throughout, $\operatorname{proj}_{Q}(z)$ denotes the orthogonal projection of
vector $z$ onto a unit vector $Q$ or onto the column span of an orthonormal basis
$Q$; when $Q$ is an orthonormal basis, this is
$\operatorname{proj}_{Q}(z)=QQ^\top z$.

\subsection{Usage-Conditioned Residual Structure}
\label{sec:residual-usage-geometry}

To isolate variation beyond the shared sentiment direction, we remove the
projection of each sentiment delta onto that direction. Since $B^\ell$ is
unit-normalized, the residual sentiment delta is
\begin{equation}
    R^\ell_{s,u}
    =
    \Delta^\ell_{s,u}
    -
    \big[(\Delta^\ell_{s,u})^\top B^\ell\big]B^\ell.
    \label{eq:residual-delta}
\end{equation}
We use $R^\ell_{s,u}$ to ask whether deltas with the same usage value still
form reproducible structure after matching event content and removing the shared
sentiment direction. As a baseline, we shuffle usage labels within each skeleton
and usage family. This keeps the positive--negative text pairs fixed, but breaks
the association between residuals and labels such as \emph{formal} or
\emph{casual}. We then measure whether a residual from an unseen skeleton is
closer to the centroid of its own usage value than expected under this
within-skeleton shuffle. Positive margins indicate usage-conditioned residual
structure beyond the shared sentiment direction.

Table~\ref{tab:shared-residual-summary} also reports projection share, residual
ratio, and usage decoding. Projection share is the fraction of squared delta
norm captured by $B^\ell$, residual ratio is
$\|R^\ell_{s,u}\|/\|\Delta^\ell_{s,u}\|$, and usage-decoding margin is
nearest-centroid test accuracy minus the 95th-percentile accuracy from the
within-skeleton label-shuffle protocol.

\subsection{Results}
\label{sec:shared-residual-results}

As shown in Table~\ref{tab:shared-residual-summary}, held-out tone and audience
mean sentiment deltas align closely with the training-set sentiment direction
(median cosine 0.953--0.975), showing a robust shared sentiment component.
However, this direction captures only 0.172--0.301 of the sample-level squared
norm, while a large residual remains (0.833--0.909 of the original norm). This
granularity distinction matters: alignment is measured on aggregated
family-level mean deltas, whereas projection share and residual ratio are
computed over individual test deltas. The residual component is not
unstructured noise: tone and audience decoding margins remain positive relative
to within-skeleton label shuffles (0.114--0.356 for tone; 0.049--0.326 for
audience). Thus, sentiment deltas contain both a strong shared direction and a
usage-conditioned residual component.

Appendix~\ref{app:residual-geometry-details} reports the corresponding
layer-wise shared-direction alignment and residual usage decoding curves in
Figures~\ref{fig:backbone-alignment-layerwise} and
\ref{fig:residual-usage-layerwise}. The residual usage signal is not tied to a
single pooling choice: mean pooling becomes more separable in later layers,
whereas last-token pooling is already separable in early-to-middle layers and
then remains flatter, with a sharper late-layer drop for Gemma than for Llama or
Mistral.
\begin{table*}[t]
    \centering
    \small
    \setlength{\tabcolsep}{2.4pt}
    \begin{tabular}{@{}llrrrrrrr@{}}
    \toprule
    Model & Usage Family & Rank-1 & Rank-2 & SelfCap--Shuf95 & GapCap--Shuf95 & Rule Pass & Value Pass & Diag. Margin \\
    \midrule
    Gemma & Audience & 0.677 & 0.901 & 0.463 & 0.389 & 0.964 & 1.000 & 0.605 \\
    Gemma & Tone & 0.757 & 0.918 & 0.591 & 0.538 & 1.000 & 1.000 & 0.433 \\
    Llama & Audience & 0.666 & 0.920 & 0.536 & 0.423 & 1.000 & 1.000 & 0.641 \\
    Llama & Tone & 0.818 & 0.948 & 0.652 & 0.585 & 1.000 & 1.000 & 0.333 \\
    Mistral & Audience & 0.677 & 0.917 & 0.476 & 0.357 & 1.000 & 1.000 & 0.578 \\
    Mistral & Tone & 0.768 & 0.926 & 0.587 & 0.526 & 1.000 & 1.000 & 0.385 \\
    \bottomrule
    \end{tabular}
    \caption{Compactness, cross-split stability, and value-level reproducibility of
    last-token residual subspaces. Rank-1 and Rank-2 are median fractions of centered
    value-mean energy captured by rank-1 and rank-2 approximations; higher values
    indicate more compact subspaces. SelfCap--Shuf95 is the median across layers of
    same-family cross-split capture above the 95th percentile shuffle baseline.
    GapCap--Shuf95 is the corresponding margin for same-family capture over
    other-family capture. Rule Pass is the fraction of layers passing the full
    value-level rule, and Value Pass is the median fraction of one-versus-rest value
    directions above the shuffle threshold. Diag. Margin is the median
    train--test diagonal-dominance margin; positive values indicate that matching
    usage-value directions are closer than mismatched directions.}
    \label{tab:subspace-value-results}
    \end{table*}
    
\section{Usage-Conditioned Residual Subspaces}
\label{sec:residual-subspaces}

Section~\ref{sec:shared-residual-geometry} showed that, after removing the
shared sentiment direction, tone and audience remain decodable from residual
sentiment deltas. This section tests whether that residual usage signal has
stable low-dimensional geometry. We use last-token residuals as the primary
setting for the subspace and erasure analyses, because this setting matches the
generation-time intervention and yields positive residual usage margins in
Table~\ref{tab:shared-residual-summary}. For each model and layer, we estimate
separate tone and audience residual subspaces. We then test two properties:
whether these subspaces reproduce when estimated independently from disjoint
training and test skeleton sets, and whether they contain reproducible
directions for individual usage values.

\subsection{Tone and Audience Subspace Stability}
\label{sec:tone-audience-subspace-stability}

We first test whether residual usage variation forms compact subspaces that
reproduce across unseen skeletons.
For each model, layer, and usage family in the last-token setting
$f \in \{\text{tone}, \text{audience}\}$, we estimate a residual subspace from
training skeletons. Let $\mathcal{V}_f$ be the four usage values belonging to
$f$. When the usage family is fixed, we write its value as
$v \in \mathcal{V}_f$ rather than the global usage value $u$; in this context
$R^\ell_{s,v}$ denotes the residual $R^\ell_{s,u}$ for the corresponding
family-specific usage value. For each value $v$, we average the residual
sentiment deltas over training skeletons:
\begin{equation}
    \mu^\ell_{f,v}
    =
    \frac{1}{|\mathcal{S}_{\mathrm{train}}|}
    \sum_{s \in \mathcal{S}_{\mathrm{train}}}
    R^\ell_{s,v}.
\end{equation}
We center the four value means, stack them into a value-by-dimension matrix, and
let $S_f^\ell$ be the rank-3 orthonormal basis spanned by its top
right-singular vectors. With four centered values, the rank is at most 3. We
also report rank-1 and rank-2 energy as compactness measures.

We evaluate stability with energy-weighted cross-split capture rather than an
unweighted basis-overlap score, because the singular-vector directions can
carry very different amounts of centered value-mean energy. In brief, we ask how
much centered train value-mean energy is captured by the independently estimated
test subspace, and vice versa, then average the two directions. Capture for the
other usage family replaces the matching-family subspace with the independently
estimated subspace for the other family. The capture gap is computed per layer
as same usage family capture minus other usage family capture. Train and test
subspaces are estimated independently; the capture statistic is used only for
evaluation and is not used to select layers, ranks, or model settings. We
compare same usage family capture and the capture gap with 500 within-skeleton
usage-label shuffles under the same estimation protocol.

Table~\ref{tab:subspace-value-results} first supports compactness and
cross-split stability in the last-token residual setting. Rank-1 captures
0.666--0.818 of the centered value-mean energy, and rank-2 captures
0.901--0.948. Same usage family energy-weighted capture exceeds the shuffle
baseline for every model and usage family, and the gap between same usage
family capture and other usage family capture is positive relative to the
shuffle baseline in all six cases. We interpret these subspaces as stable and
compact summaries of residual usage structure.

\subsection{Usage-Value Directions}
\label{sec:usage-value-directions}

A stable tone or audience subspace should also contain reproducible directions
for individual values such as \emph{formal} or \emph{casual}. For each value, we
compare its residual against the average residual of the other same-family values
within the same skeleton, average these one-versus-rest offsets separately over
train and test skeletons, and compare the resulting value directions against the
same shuffle baseline. We also check two pre-specified contrasts per usage
family:
\emph{formal}--\emph{casual} and \emph{enthusiastic}--\emph{restrained} for
tone, and \emph{specialist-facing}--\emph{general-public-facing} and
\emph{child-friendly}--\emph{peer-facing} for audience. Finally, a diagonal
dominance check asks whether each train value direction is closest to the
matching test value direction. Together, these checks test value separability,
pre-specified contrast reproducibility, and train--test matching.

The value-level rule combines these three checks. For a given model and usage
family, a layer passes if at least three of four one-versus-rest directions
exceed the shuffle 95th percentile, at least one of the two pre-specified
contrasts exceeds the shuffle 95th percentile, and the median diagonal-dominance
margin is positive. A model and usage family pair passes if at least half of its
layers pass; no layer is selected from this analysis.

Table~\ref{tab:subspace-value-results} shows a strong and consistent value-level
signal under last-token pooling. Every model and usage family pair passes the
pre-specified rule, and the rule succeeds in 183 of 184 layers. The component
checks follow the same pattern: one-versus-rest value directions,
pre-specified contrasts, and diagonal matching are almost always above their
respective baselines. The weakest case is Gemma audience, which still passes in
0.964 of layers, well above the pre-specified 0.5 threshold. Thus, the stable
residual subspaces in Section~\ref{sec:tone-audience-subspace-stability} are
grounded in reproducible usage-value directions, rather than only aggregate
subspace capture.

\section{Residual Subspace Erasure}
\label{sec:targeted-residual-erasure}

Section~\ref{sec:residual-subspaces} summarized residual usage structure with
stable tone and audience subspaces. We now test whether these subspaces carry
held-out usage information by erasing each learned subspace from test residuals
and comparing the metric drop with random and label-shuffled subspace controls.

\subsection{Erasure Protocol}
\label{sec:erasure-protocol}

For each model, layer, and usage family in the last-token setting, we reuse the
sentiment deltas and residuals $R^\ell_{s,u}$ from Eq.~\ref{eq:residual-delta}.
For a target usage family $f \in \{\text{tone}, \text{audience}\}$, we use the
rank-3 orthonormal basis $S_f^\ell$ estimated from the centered train
usage-value means. The erasure intervention removes only the target residual
subspace while preserving the projection onto the shared sentiment direction:
\begin{equation}
    \widetilde{\Delta}^{\ell,f}_{s,u}
    =
    \operatorname{proj}_{B^\ell}(\Delta^\ell_{s,u})
    +
    R^\ell_{s,u}
    -
    \operatorname{proj}_{S_f^\ell}(R^\ell_{s,u}).
\end{equation}
This design tests whether the learned subspace carries usage information beyond
the shared sentiment direction, rather than merely reducing the whole sentiment
delta.

We compare target erasure with no erasure, other-family erasure, random rank-3
residual erasure, and label-shuffled same-family rank-3 erasure. Evaluation uses
retention relative to no erasure for centroid separation, contrast consistency,
and decoding, then averages the three retentions per layer. Positive
target-minus-control reduction gaps mean target erasure reduces the target usage
metrics more than the random or shuffled controls. The target/non-target
selectivity gap is defined analogously by comparing the target-family reduction
with the non-target-family reduction under the same target erasure. A layer
passes if target erasure reduces the target usage metrics more than the random
and shuffled controls while retaining at least 0.95 of the shared-direction
projection. A model and usage family pass if this rule holds in at least half of
layers and shared-direction projection retention exceeds 0.95 in at least 95\%
of layers. Results are aggregated over all layers rather than selected from a
best layer.

\subsection{Results}
\label{sec:erasure-results}

\begin{table}[t]
    \centering
    \scriptsize
    \setlength{\tabcolsep}{3pt}
    \begin{tabular}{@{}llrrrrr@{}}
    \toprule
    Model & Usage & Target & Rand. & Shuf. & Gaps & Sel. \\
    \midrule
    Llama$_{32}$ & Tone & 0.456 & 1.000 & 0.972 & 0.545/0.502 & 0.540 \\
    Llama$_{32}$ & Audience & 0.541 & 1.000 & 0.989 & 0.459/0.439 & 0.399 \\
    Mistral$_{32}$ & Tone & 0.539 & 1.000 & 0.980 & 0.461/0.447 & 0.446 \\
    Mistral$_{32}$ & Audience & 0.644 & 1.000 & 0.991 & 0.359/0.344 & 0.326 \\
    Gemma$_{28}$ & Tone & 0.500 & 1.000 & 0.982 & 0.499/0.477 & 0.474 \\
    Gemma$_{28}$ & Audience & 0.614 & 1.000 & 0.993 & 0.386/0.378 & 0.365 \\
    \bottomrule
    \end{tabular}
    \caption{Targeted residual erasure results under the last-token
    setting. Subscripts indicate the number of transformer layers evaluated.
    Target, Rand., and Shuf. are median across-layer retentions of the averaged
    test-split usage metrics after target, random rank-3 residual, and
    label-shuffled same-family rank-3 erasure, respectively. Gaps compare
    reductions, computed as
    control-retention minus target-retention for random/shuffled controls;
    positive values mean target erasure causes a larger reduction. Sel. is the
    target/non-target reduction gap.}
    \label{tab:targeted-erasure-results}
    \end{table}

Table~\ref{tab:targeted-erasure-results} shows that erasing the learned target
subspace selectively weakens held-out usage metrics. All six model and usage
family pairs satisfy the pre-specified rule: target erasure leaves only
0.456--0.644 median retention, while random erasure stays near 1.0 and shuffled
same-family erasure reduces the metrics less than the true target subspace.

The auxiliary checks follow the same pattern. The shared-direction projection
retention criterion is satisfied for all six cases, and target erasure reduces
metrics for the target usage family more than for the non-target usage family.
These comparisons indicate that the estimated residual subspaces carry usage
information used by the test-split metrics; Appendix~\ref{app:erasure-details}
reports layer-level results.


\section{Generation-Time Steering}
\label{sec:generation-diagnostic}

We next test whether residualized activation-space tone components can steer
model behavior. If these components carry controllable usage information,
injecting them during generation should move outputs toward a target tone while
preserving the requested sentiment polarity. Because activation steering must
be evaluated for effects on general capabilities and text quality, and combining
steering vectors for multiple behaviors can be ineffective~\citep{
rimsky2024steering,vanderweij2024extending}, we use a constrained setting where
the prompt specifies a source tone, event, and outcome, while the activation
intervention supplies the target tone. We evaluate tone as the primary generation
setting and audience steering in a separate held-out setting.

\begin{table}[t]
\centering
\footnotesize
\setlength{\tabcolsep}{2pt}
\begin{tabular}{@{}lcccc@{}}
\toprule
Source $\rightarrow$ target & Src. & Tone W/T/L & Pol. & Art. \\
\midrule
formal $\rightarrow$ casual & 40/40 & 158/0/2 & 40/40 & 1 \\
restr. $\rightarrow$ enth. & 40/40 & 128/29/3 & 39/40 & 0 \\
casual $\rightarrow$ formal & 40/40 & 160/0/0 & 40/40 & 0 \\
enth. $\rightarrow$ restr. & 32/40 & 118/1/9 & 31/32 & 0 \\
\bottomrule
\end{tabular}
\caption{Llama generation-time tone steering under the fixed L20,
$\beta=50$ setting. Src. reports source-gate passes among 40 prompts:
no-intervention outputs judged closer to the prompted source tone than to the
target tone. Tone W/T/L reports wins/ties/losses for target injection in blind
GPT-4o pairwise comparisons, pooled over four controls: no intervention, a
random residual direction, a wrong target tone, and the opposite target tone.
Pol. reports polarity preservation among retained target-injected outputs. Art.
counts obvious formatting artifacts or tone-label leakage among target-injected
outputs. Enth. abbreviates enthusiastic and restr. abbreviates restrained.}
\label{tab:generation-diagnostic-main}
\end{table}

\subsection{Setup}
\label{sec:generation-intervention}

For the main-text generation experiment, we use Llama. Let
$\mathcal{V}_{\mathrm{tone}}$ be the four tone values and
$y \in \{+,-\}$ be the sentiment frame. For each layer $\ell$ and
tone value $v \in \mathcal{V}_{\mathrm{tone}}$, we estimate a last-token
activation-space tone component $A^\ell_v$ from train skeletons by averaging,
over skeletons and sentiment frames, the difference between the representation
for tone $v$ and the mean representation over all four tones for the same
skeleton and sentiment frame. We then remove its projection onto the shared
sentiment direction and normalize:
\begin{equation}
    a^\ell_v
    =
    \frac{A^\ell_v-\operatorname{proj}_{B^\ell}(A^\ell_v)}
    {\|A^\ell_v-\operatorname{proj}_{B^\ell}(A^\ell_v)\|}.
\end{equation}
During generation, target-tone injection adds this direction to the unpooled
hidden state $h^\ell_t$ while projecting the update away from the shared
sentiment direction:
\begin{equation}
    h^{\ell,\mathrm{inj}}_t
    =
    h^\ell_t
    +
    \operatorname{proj}_{(B^\ell)^\perp}
    \left(\beta\,\mathrm{RMS}(h^\ell_t)\,a^\ell_v\right).
\end{equation}
Here $\mathrm{RMS}(z)=\|z\|_2/\sqrt{D}$ for hidden size $D$. The expanded test
uses layer 20 and $\beta=50$, with the steering strength selected on the
development split.

Prompts list the four tone values with short definitions and ask for one English
sentence in a specified source tone. They then give only an event and a positive
or negative outcome frame, and instruct the model to keep the event and outcome
unchanged without naming the tone label. We evaluate four source--target pairs:
formal $\rightarrow$ casual, casual $\rightarrow$ formal, restrained
$\rightarrow$ enthusiastic, and enthusiastic $\rightarrow$ restrained. A GPT-4o
source gate keeps only prompts whose no-intervention output is closer to the
source tone than to the target tone, preventing the evaluation from counting
prompts where the baseline already lacks the requested source tone.

For each retained prompt, we compare target injection with four controls: no
intervention, a random $B^\ell$-orthogonal residual direction, a wrong target
tone, and the opposite target tone. A blind GPT-4o judge sees paired outputs in
random order and chooses which output better matches the target tone, or a tie.
Separately, a binary GPT-4o check asks whether the target-injected output still
preserves the requested positive or negative outcome. We report artifacts and
tone-label leakage as sanity checks, but the main quantities are target-tone
wins and polarity preservation.

\subsection{Tone-Steering Results}
\label{sec:generation-results}

The source gate confirms that retained prompts start from source-matched
outputs: 152 of 160 no-intervention outputs are judged closer to the source tone
than to the target tone, with no target-tone matches and only eight neither
judgments. Table~\ref{tab:generation-diagnostic-main} shows that
target injection shifts outputs toward the target tone across all four
source--target directions, including against random, wrong-target, and
opposite-target controls. The largest margins occur for casual $\rightarrow$
formal and formal $\rightarrow$ casual, where the pooled target-tone scores are
160/0/0 and 158/0/2. The enthusiastic target produces more ties, but still has
128 wins and only 3 losses. These controls make the diagnostic stricter: the
random residual direction controls for adding activation energy, while the wrong
and opposite tone controls test whether the effect is target-specific rather
than a generic style shift.

The polarity check shows that tone steering usually preserves the requested
sentiment polarity: 150 of 152 target-injected outputs (98.7\%) keep the
requested positive or negative polarity. Formatting artifacts are rare: one
target-injected output contains an explanatory preface. This supports a
target-specific tone shift for Llama under the fixed intervention setting.
Section~\ref{sec:generation-robustness} further tests judge robustness and
transfer beyond the main tone-steering evaluation.
Appendix~\ref{app:generation-diagnostic-details} gives representative
generations.

\subsection{Robustness and Transfer}
\label{sec:generation-robustness}

We first test whether the main tone-steering result is sensitive to the choice
of model judge. We re-evaluate the same 608 retained comparisons with Claude
Sonnet 5 (\texttt{claude-sonnet-5})~\citep{anthropic2026sonnet5}
as the judge. Target injection obtains 569 wins, 33 ties, and 6 losses (93.6\% wins
when ties remain in the denominator). The prompts, generated outputs, source
gate, and intervention conditions are held fixed; only the judge changes.

We next test whether the Llama layer-20 formal--casual direction transfers to
controlled human-written text. A human writer produced 40 sentences from 10
held-out skeletons, with one formal and one casual version for each positive and
negative outcome. None of these sentences is used to estimate the direction.
Without retraining, the learned direction correctly orders 18/20 matched
formal--casual pairs (90.0\%), compared with 9.96/20 on average for matched
random directions (empirical $p=.0045$).

Finally, we apply the intervention to one audience contrast on 20 held-out
specialist-facing prompts. For specialist-facing $\rightarrow$ child-friendly
steering at layer 24 and $\beta=40$, the target audience direction is preferred
to a matched random residual direction in 19 of 20 blind Claude Sonnet 5 comparisons and
to no intervention in 18 of 20 comparisons. The separate preservation audit
finds no outcome-preservation loss, while 5 of 20 target-injected outputs
contain surface-form artifacts. Full breakdowns for these evaluations appear
in Appendix~\ref{app:generation-diagnostic-details}.

\vspace{-0.2em}
\section{Conclusion}

We examined whether sentiment representations in LLMs are fully explained by a
shared sentiment direction, or whether usage modulates structure beyond that
direction. Using controlled positive--negative pairs that hold event content
fixed while varying tone and audience adaptation, we found a robust shared
sentiment direction across Llama, Mistral, and Gemma. After removing this
direction, however, the residuals remain organized by usage rather than
collapsing to noise. This residual structure is compact, and its subspaces
reproduce between disjoint training and test skeleton sets. Targeted erasure
weakens held-out usage metrics more than random or label-shuffled controls. In
generation, steering along residualized tone components
shifts Llama outputs toward a target tone while preserving the requested
sentiment polarity in 98.7\% of evaluated prompts. Independent judging
corroborates this result, while held-out human-written and audience evaluations
show transfer beyond the main tone evaluation. Taken together, these results
show that a shared sentiment direction does not fully characterize sentiment
representations in our controlled setting: the remaining variation is
systematically structured by communicative usage.

\section*{Limitations}

Our analysis is limited to a controlled synthetic setting with 221 complete
event skeletons and three instruction-tuned models at the 7B/8B scale. The
generation-time steering experiments are limited to Llama and model-based
judgments; they test directional steering under fixed settings rather than a
complete editing method.

\section*{Ethical Considerations and Potential Risks}

This work uses controlled synthetic text and does not use private user data or
study deployment in user-facing systems. Because the paper includes
generation-time tone and audience steering, similar techniques could be used to alter the
style or affective framing of generated text. While such control may support
benign editing, it could also be misused to make generated messages more
persuasive, emotionally manipulative, or misleading, especially in sensitive
settings such as mental health support. Our experiments are diagnostic and
controlled, and we do not claim deployment readiness without further safety
evaluation.

\section*{Acknowledgments}
This work is supported in part by the National Science Foundation (NSF) grant IIS-2451436 and Commonwealth Cyber Initiative grant HC-2Q26-032.

\bibliography{custom}

@article{grattafiori2024llama,
  title={The {Llama} 3 Herd of Models},
  author={Grattafiori, Aaron and Dubey, Abhimanyu and Jauhri, Abhinav and Pandey, Abhinav and Kadian, Abhishek and Al-Dahle, Ahmad and Letman, Aiesha and Mathur, Akhil and Schelten, Alan and Vaughan, Alex and others},
  journal={arXiv preprint arXiv:2407.21783},
  year={2024},
  url={https://arxiv.org/abs/2407.21783}
}

@article{team2024gemma,
  title={{Gemma}: Open Models Based on {Gemini} Research and Technology},
  author={Team, Gemma and Mesnard, Thomas and Hardin, Cassidy and Dadashi, Robert and Bhupatiraju, Surya and Pathak, Shreya and Sifre, Laurent and Rivi{\`e}re, Morgane and Kale, Mihir Sanjay and Love, Juliette and others},
  journal={arXiv preprint arXiv:2403.08295},
  year={2024},
  url={https://arxiv.org/abs/2403.08295}
}

@misc{jiang2023mistral7b,
      title={{Mistral} {7B}}, 
      author={Albert Q. Jiang and Alexandre Sablayrolles and Arthur Mensch and Chris Bamford and Devendra Singh Chaplot and Diego de las Casas and Florian Bressand and Gianna Lengyel and Guillaume Lample and Lucile Saulnier and Lélio Renard Lavaud and Marie-Anne Lachaux and Pierre Stock and Teven Le Scao and Thibaut Lavril and Thomas Wang and Timothée Lacroix and William El Sayed},
      year={2023},
      eprint={2310.06825},
      archivePrefix={arXiv},
      primaryClass={cs.CL},
      url={https://arxiv.org/abs/2310.06825}, 
}

@article{zou2023representation,
  title={Representation Engineering: A Top-Down Approach to {AI} Transparency},
  author={Zou, Andy and Phan, Long and Chen, Sarah and Campbell, James and Guo, Phillip and Ren, Richard and Pan, Alexander and Yin, Xuwang and Mazeika, Mantas and Dombrowski, Ann-Kathrin and Goel, Shashwat and Li, Nathaniel and Byun, Michael J. and Wang, Zifan and Mallen, Alex and Basart, Steven and Koyejo, Sanmi and Song, Dawn and Fredrikson, Matt and Kolter, J. Zico and Hendrycks, Dan},
  journal={arXiv preprint arXiv:2310.01405},
  year={2023},
  url={https://arxiv.org/abs/2310.01405}
}

@article{brown2020language,
  title={Language Models are Few-Shot Learners},
  author={Brown, Tom and Mann, Benjamin and Ryder, Nick and Subbiah, Melanie and Kaplan, Jared D and Dhariwal, Prafulla and Neelakantan, Arvind and Shyam, Pranav and Sastry, Girish and Askell, Amanda and others},
  journal={Advances in Neural Information Processing Systems},
  volume={33},
  pages={1877--1901},
  year={2020},
  url={https://proceedings.neurips.cc/paper/2020/hash/1457c0d6bfcb4967418bfb8ac142f64a-Abstract.html}
}

@misc{radford2019language,
  author       = {Radford, Alec and Wu, Jeffrey and Child, Rewon and
                  Luan, David and Amodei, Dario and Sutskever, Ilya},
  title        = {Language Models are Unsupervised Multitask Learners},
  year         = {2019},
  howpublished = {OpenAI technical report},
  url          = {https://cdn.openai.com/better-language-models/language_models_are_unsupervised_multitask_learners.pdf}
}

@inproceedings{yang2023towards,
    title = "Towards Interpretable Mental Health Analysis with Large Language Models",
    author = "Yang, Kailai  and
      Ji, Shaoxiong  and
      Zhang, Tianlin  and
      Xie, Qianqian  and
      Kuang, Ziyan  and
      Ananiadou, Sophia",
    editor = "Bouamor, Houda  and
      Pino, Juan  and
      Bali, Kalika",
    booktitle = "Proceedings of the 2023 Conference on Empirical Methods in Natural Language Processing",
    month = dec,
    year = "2023",
    address = "Singapore",
    publisher = "Association for Computational Linguistics",
    url = "https://aclanthology.org/2023.emnlp-main.370/",
    doi = "10.18653/v1/2023.emnlp-main.370",
    pages = "6056--6077"
}

@article{hu2025beyond,
  title={Beyond Empathy: Integrating Diagnostic and Therapeutic Reasoning with Large Language Models for Mental Health Counseling},
  author={Hu, He and Zhou, Yucheng and Si, Juzheng and Wang, Qianning and Zhang, Hengheng and Ren, Fuji and Ma, Fei and Cui, Laizhong and Tian, Qi},
  journal={arXiv preprint arXiv:2505.15715},
  year={2025},
  url={https://arxiv.org/abs/2505.15715}
}

@inproceedings{hollinsworth-etal-2024-language,
    title = "Language Models Linearly Represent Sentiment",
    author = "Tigges, Curt  and
      Hollinsworth, Oskar J.  and
      Geiger, Atticus  and
      Nanda, Neel",
    editor = "Belinkov, Yonatan  and
      Kim, Najoung  and
      Jumelet, Jaap  and
      Mohebbi, Hosein  and
      Mueller, Aaron  and
      Chen, Hanjie",
    booktitle = "Proceedings of the 7th BlackboxNLP Workshop: Analyzing and Interpreting Neural Networks for NLP",
    month = nov,
    year = "2024",
    address = "Miami, Florida, US",
    publisher = "Association for Computational Linguistics",
    url = "https://aclanthology.org/2024.blackboxnlp-1.5/",
    doi = "10.18653/v1/2024.blackboxnlp-1.5",
    pages = "58--87"
}

@InProceedings{park2023linear,
  title = 	 {The Linear Representation Hypothesis and the Geometry of Large Language Models},
  author =       {Park, Kiho and Choe, Yo Joong and Veitch, Victor},
  booktitle = 	 {Proceedings of the 41st International Conference on Machine Learning},
  pages = 	 {39643--39666},
  year = 	 {2024},
  editor = 	 {Salakhutdinov, Ruslan and Kolter, Zico and Heller, Katherine and Weller, Adrian and Oliver, Nuria and Scarlett, Jonathan and Berkenkamp, Felix},
  volume = 	 {235},
  series = 	 {Proceedings of Machine Learning Research},
  month = 	 {21--27 Jul},
  publisher =    {PMLR},
  url = 	 {https://proceedings.mlr.press/v235/park24c.html}
  }

@InProceedings{jiang2024origins,
  title = 	 {On the Origins of Linear Representations in Large Language Models},
  author =       {Jiang, Yibo and Rajendran, Goutham and Ravikumar, Pradeep Kumar and Aragam, Bryon and Veitch, Victor},
  booktitle = 	 {Proceedings of the 41st International Conference on Machine Learning},
  pages = 	 {21879--21911},
  year = 	 {2024},
  editor = 	 {Salakhutdinov, Ruslan and Kolter, Zico and Heller, Katherine and Weller, Adrian and Oliver, Nuria and Scarlett, Jonathan and Berkenkamp, Felix},
  volume = 	 {235},
  series = 	 {Proceedings of Machine Learning Research},
  month = 	 {21--27 Jul},
  publisher =    {PMLR},
  url = 	 {https://proceedings.mlr.press/v235/jiang24d.html}
}

@inproceedings{palma2025llamas,
    title = "{LL}a{MA}s Have Feelings Too: Unveiling Sentiment and Emotion Representations in {LL}a{MA} Models Through Probing",
    author = "Di Palma, Dario  and
      De Bellis, Alessandro  and
      Servedio, Giovanni  and
      Anelli, Vito Walter  and
      Narducci, Fedelucio  and
      Di Noia, Tommaso",
    editor = "Che, Wanxiang  and
      Nabende, Joyce  and
      Shutova, Ekaterina  and
      Pilehvar, Mohammad Taher",
    booktitle = "Proceedings of the 63rd Annual Meeting of the Association for Computational Linguistics (Volume 1: Long Papers)",
    month = jul,
    year = "2025",
    address = "Vienna, Austria",
    publisher = "Association for Computational Linguistics",
    url = "https://aclanthology.org/2025.acl-long.306/",
    doi = "10.18653/v1/2025.acl-long.306",
    pages = "6124--6142",
    ISBN = "979-8-89176-251-0"
}

@article{choi2026latent,
  title={Latent Structure of Affective Representations in Large Language Models},
  author={Choi, Benjamin J. and Weber, Melanie},
  journal={arXiv preprint arXiv:2604.07382},
  year={2026},
  url={https://arxiv.org/abs/2604.07382}
}

@article{sun2026valence,
  title={Valence-Arousal Subspace in {LLM}s: Circular Emotion Geometry and Multi-Behavioral Control},
  author={Sun, Lihao and Yan, Lewen and Lu, Xiaoya and Lee, Andrew and Zhang, Jie and Shao, Jing},
  journal={arXiv preprint arXiv:2604.03147},
  year={2026},
  url={https://arxiv.org/abs/2604.03147}
}

@inproceedings{
pahde2022navigating,
title={Navigating Neural Space: Revisiting Concept Activation Vectors to Overcome Directional Divergence},
author={Frederik Pahde and Maximilian Dreyer and Moritz Weckbecker and Leander Weber and Christopher J. Anders and Thomas Wiegand and Wojciech Samek and Sebastian Lapuschkin},
booktitle={The Thirteenth International Conference on Learning Representations},
year={2025},
url={https://openreview.net/forum?id=Q95MaWfF4e}
}

@incollection{clark1982audience,
  author    = {Clark, Herbert H. and Murphy, Gregory L.},
  title     = {Audience Design in Meaning and Reference},
  booktitle = {Language and Comprehension},
  editor    = {Le Ny, Jean-Fran{\c{c}}ois and Kintsch, Walter},
  series    = {Advances in Psychology},
  volume    = {9},
  pages     = {287--299},
  publisher = {North-Holland},
  address   = {Amsterdam},
  year      = {1982},
  doi       = {10.1016/S0166-4115(09)60059-5}
}

@article{bell1984language,
  author  = {Bell, Allan},
  title   = {Language Style as Audience Design},
  journal = {Language in Society},
  volume  = {13},
  number  = {2},
  pages   = {145--204},
  year    = {1984},
  doi     = {10.1017/S004740450001037X}
}

@article{arditi2024refusal,
  title={Refusal in Language Models Is Mediated by a Single Direction},
  author={Arditi, Andy and Obeso, Oscar and Syed, Aaquib and Paleka, Daniel and Panickssery, Nina and Gurnee, Wes and Nanda, Neel},
  journal={Advances in Neural Information Processing Systems},
  volume={37},
  pages={136037--136083},
  year={2024},
  doi={10.52202/079017-4322},
  url={https://proceedings.neurips.cc/paper_files/paper/2024/hash/f545448535dfde4f9786555403ab7c49-Abstract-Conference.html}
}

@inproceedings{mikolov2013linguistic,
  title={Linguistic Regularities in Continuous Space Word Representations},
  author={Mikolov, Tomas and Yih, Wen-tau and Zweig, Geoffrey},
  booktitle={Proceedings of the 2013 Conference of the North American Chapter of the Association for Computational Linguistics: Human Language Technologies},
  pages={746--751},
  year={2013},
  address={Atlanta, Georgia},
  publisher={Association for Computational Linguistics},
  url={https://aclanthology.org/N13-1090/}
}

@article{li2023inference,
  title={Inference-Time Intervention: Eliciting Truthful Answers from a Language Model},
  author={Li, Kenneth and Patel, Oam and Vi{\'e}gas, Fernanda and Pfister, Hanspeter and Wattenberg, Martin},
  journal={Advances in Neural Information Processing Systems},
  volume={36},
  pages={41451--41530},
  year={2023},
  doi={10.52202/075280-1797},
  url={https://proceedings.neurips.cc/paper_files/paper/2023/hash/81b8390039b7302c909cb769f8b6cd93-Abstract-Conference.html}
}

@article{marks2023geometry,
  title={The Geometry of Truth: Emergent Linear Structure in Large Language Model Representations of True/False Datasets},
  author={Marks, Samuel and Tegmark, Max},
  journal={arXiv preprint arXiv:2310.06824},
  year={2023},
  url={https://arxiv.org/abs/2310.06824}
}

@inproceedings{
zhao2025beyond,
title={Beyond Single Concept Vector: Modeling Concept Subspace in {LLM}s with Gaussian Distribution},
author={Haiyan Zhao and Heng Zhao and Bo Shen and Ali Payani and Fan Yang and Mengnan Du},
booktitle={The Thirteenth International Conference on Learning Representations},
year={2025},
url={https://openreview.net/forum?id=CvttyK4XzV}
}

@misc{openai2024gpt4o,
  author       = {{OpenAI}},
  title        = {Hello {GPT-4o}},
  year         = {2024},
  howpublished = {OpenAI},
  url          = {https://openai.com/index/hello-gpt-4o/},
  note         = {Accessed: 2026-08-30}
}

@misc{openai2024gpt4omini,
  author       = {{OpenAI}},
  title        = {{GPT-4o mini}: advancing cost-efficient intelligence},
  year         = {2024},
  howpublished = {OpenAI},
  url          = {https://openai.com/index/gpt-4o-mini-advancing-cost-efficient-intelligence/},
  note         = {Accessed: 2026-08-30}
}

@misc{anthropic2026sonnet5,
  author       = {{Anthropic}},
  title        = {Introducing {Claude Sonnet 5}},
  year         = {2026},
  month        = jun,
  url          = {https://www.anthropic.com/news/claude-sonnet-5},
  note         = {Accessed: 2026-08-30}
}

@article{chen2025persona,
  title={Persona Vectors: Monitoring and Controlling Character Traits in Language Models},
  author={Chen, Runjin and Arditi, Andy and Sleight, Henry and Evans, Owain and Lindsey, Jack},
  journal={arXiv preprint arXiv:2507.21509},
  year={2025},
  url={https://arxiv.org/abs/2507.21509}
}

@article{turner2023steering,
  title={Steering Language Models with Activation Engineering},
  author={Turner, Alexander Matt and Thiergart, Lisa and Leech, Gavin and Udell, David and Vazquez, Juan J and Mini, Ulisse and MacDiarmid, Monte},
  journal={arXiv preprint arXiv:2308.10248},
  year={2023},
  url={https://arxiv.org/abs/2308.10248}
}

@InProceedings{liu2023incontext,
  title = 	 {In-context Vectors: Making In Context Learning More Effective and Controllable Through Latent Space Steering},
  author =       {Liu, Sheng and Ye, Haotian and Xing, Lei and Zou, James Y.},
  booktitle = 	 {Proceedings of the 41st International Conference on Machine Learning},
  pages = 	 {32287--32307},
  year = 	 {2024},
  editor = 	 {Salakhutdinov, Ruslan and Kolter, Zico and Heller, Katherine and Weller, Adrian and Oliver, Nuria and Scarlett, Jonathan and Berkenkamp, Felix},
  volume = 	 {235},
  series = 	 {Proceedings of Machine Learning Research},
  month = 	 {21--27 Jul},
  publisher =    {PMLR},
  url = 	 {https://proceedings.mlr.press/v235/liu24bx.html}
}

@article{vanderweij2024extending,
  title={Extending Activation Steering to Broad Skills and Multiple Behaviours},
  author={van der Weij, Teun and Poesio, Massimo and Schoots, Nandi},
  journal={arXiv preprint arXiv:2403.05767},
  year={2024},
  url={https://arxiv.org/abs/2403.05767}
}

@inproceedings{rimsky2024steering,
  title     = {Steering Llama 2 via Contrastive Activation Addition},
  author    = {Rimsky, Nina and Gabrieli, Nick and Schulz, Julian and Tong, Meg and Hubinger, Evan and Turner, Alexander},
  booktitle = {Proceedings of the 62nd Annual Meeting of the Association for Computational Linguistics (Volume 1: Long Papers)},
  year      = {2024},
  pages     = {15504--15522},
  address   = {Bangkok, Thailand},
  publisher = {Association for Computational Linguistics},
  url       = {https://aclanthology.org/2024.acl-long.828/},
  doi       = {10.18653/v1/2024.acl-long.828}
}

@inproceedings{gabriel2024can,
    title = "Can {AI} Relate: Testing Large Language Model Response for Mental Health Support",
    author = "Gabriel, Saadia  and
      Puri, Isha  and
      Xu, Xuhai  and
      Malgaroli, Matteo  and
      Ghassemi, Marzyeh",
    editor = "Al-Onaizan, Yaser  and
      Bansal, Mohit  and
      Chen, Yun-Nung",
    booktitle = "Findings of the Association for Computational Linguistics: EMNLP 2024",
    month = nov,
    year = "2024",
    address = "Miami, Florida, USA",
    publisher = "Association for Computational Linguistics",
    url = "https://aclanthology.org/2024.findings-emnlp.120/",
    doi = "10.18653/v1/2024.findings-emnlp.120",
    pages = "2206--2221"
}

@inproceedings{kaushik2020learning,
  title     = {Learning the Difference That Makes a Difference with Counterfactually-Augmented Data},
  author    = {Kaushik, Divyansh and Hovy, Eduard and Lipton, Zachary C.},
  booktitle = {International Conference on Learning Representations},
  year      = {2020},
  url       = {https://openreview.net/forum?id=Sklgs0NFvr}
}

@inproceedings{gardner2020evaluating,
  title     = {Evaluating Models' Local Decision Boundaries via Contrast Sets},
  author    = {Gardner, Matt and Artzi, Yoav and Basmov, Victoria and Berant, Jonathan and Bogin, Ben and Chen, Sihao and Dasigi, Pradeep and Dua, Dheeru and Elazar, Yanai and Gottumukkala, Ananth and Gupta, Nitish and Hajishirzi, Hannaneh and Ilharco, Gabriel and Khashabi, Daniel and Lin, Kevin and Liu, Jiangming and Liu, Nelson F. and Mulcaire, Phoebe and Ning, Qiang and Singh, Sameer and Smith, Noah A. and Subramanian, Sanjay and Tsarfaty, Reut and Wallace, Eric and Zhang, Ally and Zhou, Ben},
  booktitle = {Findings of the Association for Computational Linguistics: EMNLP 2020},
  year      = {2020},
  pages     = {1307--1323},
  publisher = {Association for Computational Linguistics},
  url       = {https://aclanthology.org/2020.findings-emnlp.117/},
  doi       = {10.18653/v1/2020.findings-emnlp.117}
}

@inproceedings{ficler2017controlling,
  title     = {Controlling Linguistic Style Aspects in Neural Language Generation},
  author    = {Ficler, Jessica and Goldberg, Yoav},
  booktitle = {Proceedings of the Workshop on Stylistic Variation},
  year      = {2017},
  pages     = {94--104},
  address   = {Copenhagen, Denmark},
  publisher = {Association for Computational Linguistics},
  url       = {https://aclanthology.org/W17-4912/},
  doi       = {10.18653/v1/W17-4912}
}

@article{niu2018polite,
  title     = {Polite Dialogue Generation Without Parallel Data},
  author    = {Niu, Tong and Bansal, Mohit},
  journal   = {Transactions of the Association for Computational Linguistics},
  volume    = {6},
  year      = {2018},
  pages     = {373--389},
  address   = {Cambridge, MA},
  publisher = {MIT Press},
  url       = {https://aclanthology.org/Q18-1027/},
  doi       = {10.1162/tacl_a_00027}
}

@inproceedings{moorjani2022audience,
  title     = {Audience-Centric Natural Language Generation via Style Infusion},
  author    = {Moorjani, Samraj and Krishnan, Adit and Sundaram, Hari and Maslowska, Ewa and Sankar, Aravind},
  booktitle = {Findings of the Association for Computational Linguistics: EMNLP 2022},
  year      = {2022},
  pages     = {1919--1932},
  address   = {Abu Dhabi, United Arab Emirates},
  publisher = {Association for Computational Linguistics},
  url       = {https://aclanthology.org/2022.findings-emnlp.138/},
  doi       = {10.18653/v1/2022.findings-emnlp.138}
}

@inproceedings{shu2020controllable,
  title     = {Controllable Text Generation with Focused Variation},
  author    = {Shu, Lei and Papangelis, Alexandros and Wang, Yi-Chia and Tur, Gokhan and Xu, Hu and Feizollahi, Zhaleh and Liu, Bing and Molino, Piero},
  booktitle = {Findings of the Association for Computational Linguistics: EMNLP 2020},
  year      = {2020},
  pages     = {3805--3817},
  address   = {Online},
  publisher = {Association for Computational Linguistics},
  url       = {https://aclanthology.org/2020.findings-emnlp.339/},
  doi       = {10.18653/v1/2020.findings-emnlp.339}
}

\appendix

\section{Dataset Construction Details}
\label{app:dataset-details}

This appendix gives the construction details behind the controlled paired
dataset used throughout the representation experiments. Its purpose is to make
clear how the paper obtains matched sentiment pairs while keeping the event
skeleton, usage family, and polarity labels explicit. Table~\ref{tab:dataset-details}
summarizes the resulting tone--audience dataset. The dataset contains 221 final
event skeletons. For each usage family, every skeleton is realized under four
usage values and two sentiment polarities, which gives 1768 rows per usage
family and 3536 rows in total.

\begin{table*}[t]
\centering
\small
\begin{tabular}{@{}p{0.18\linewidth}p{0.42\linewidth}p{0.16\linewidth}p{0.16\linewidth}@{}}
\toprule
Usage Family & Usage values & Construction & Rows \\
\midrule
Tone &
\emph{formal}, \emph{casual}, \emph{enthusiastic}, \emph{restrained} &
221 skeletons $\times$ 4 values $\times$ 2 polarities &
1768 \\
Audience adaptation &
\emph{specialist-facing}, \emph{general-public-facing},
\emph{child-friendly}, \emph{peer-facing} &
221 skeletons $\times$ 4 values $\times$ 2 polarities &
1768 \\
\midrule
Total & -- & 221 skeletons $\times$ 2 families $\times$ 4 values $\times$ 2 polarities &
3536 \\
\bottomrule
\end{tabular}
\caption{Controlled tone--audience dataset size and usage values. Each
usage-value/sentiment cell contains 221 examples.}
\label{tab:dataset-details}
\end{table*}

Table~\ref{tab:dataset-generation-details} summarizes the data-generation and
filtering protocol. For each skeleton, the generation code enumerates target
rows by taking every requested usage value and both sentiment frames, then asks
the model to add only the \emph{text} field while preserving the metadata. Thus,
a \emph{formal} example is not produced by a separate formula: it is the target
row whose \emph{usage\_type} is \emph{tone} and whose \emph{usage\_value} is
\emph{formal}, generated under the tone prompt definition below.

The two usage families are defined by the generation and validation prompts.
For tone, the prompt defines \emph{formal} as polished, professional, and
precise wording; \emph{casual} as conversational and relaxed wording;
\emph{enthusiastic} as energetic but factually grounded wording; and
\emph{restrained} as understated and measured wording. For audience adaptation,
the prompt defines \emph{specialist-facing} for readers with background
knowledge, \emph{general-public-facing} for ordinary adult readers,
\emph{child-friendly} through short, simple, concrete wording, and
\emph{peer-facing} for an adult peer, classmate, coworker, or familiar reader.
Audience realizations are also constrained to keep the same actor, event,
timing, polarity, and outcome.
Appendix~\ref{app:generation-prompts} lists the full prompt templates and
additional prompt constraints.

Audience values were selected to preserve the event rather than introduce new
decision content. In pilot generation, a
\emph{decision-maker-facing} value often added risks, priorities, or
next-step framing that was not part of the original skeleton, so we replaced it
with \emph{peer-facing}.

\begin{table*}[t]
\centering
\footnotesize
\begin{tabular}{@{}p{0.16\linewidth}p{0.17\linewidth}p{0.22\linewidth}p{0.34\linewidth}@{}}
\toprule
Stage & Model / Temp. & Output & Main checks \\
\midrule
Skeleton generation &
GPT-4o / 0.5 &
300 skeletons &
Sequential IDs; fixed domain schedule; neutral 8--18 word events; parallel
positive/negative frames; no usage-style labels; duplicate-event rejection. \\
Tone realization &
GPT-4o / 0.5 &
2,400 initial tone rows; 1,768 retained tone rows used here &
JSON schema; metadata preservation; 15--35 words; no label leakage, greetings,
hashtags, emojis, markdown, sign-offs; conservative block-level usage filter. \\
Audience realization &
GPT-4o / 0.4; repair 0.3 &
1,768 retained audience rows aligned to tone-complete skeletons &
JSON schema; 18--35 words; same actor/event/timing/polarity/outcome; bans
new causes, remedies, decisions, risks, priorities, budgets, and label words;
specialist-facing new-metric risk checks. \\
\bottomrule
\end{tabular}
\caption{Dataset construction protocol. Prompts are implemented as JSON
generation scripts and contain the schema and constraints summarized here; the
experiments use the retained balanced tone--audience rows.}
\label{tab:dataset-generation-details}
\end{table*}

The deterministic quality-control pass checks schema validity, duplicate keys,
cell balance, length constraints, generation artifacts, label leakage, and
forbidden new facts. The tone rows come from the conservative block-level
filtered controlled set: 300 generated skeletons produce 300 tone blocks, of
which 221 complete tone blocks are retained. The final merged tone--audience
dataset has no hard-dropped blocks under the later tone--audience QC pass. The remaining
warnings are diagnostic rather than removal criteria: some audience examples
have low lexical overlap with skeleton keywords, and some specialist-facing
examples risk adding domain-like framing such as metrics or mechanisms. These
warnings motivate treating audience adaptation as a harder usage family than
tone in the main text.

\subsection{Blind Validation Protocol}

The human validation sheets contain the base event, sentiment frame, usage
contrast and definitions, and two realizations labeled only A and B. Private
files containing the gold assignment are kept separate from the annotation
sheets. One human annotator completes all 40 tone and 40 audience items, which
are selected to balance contrast, sentiment polarity, and which usage value
appears as A. For each item, the annotator chooses one of the two assignments or marks the pair tied/unclear,
then independently records event and sentiment preservation for both texts.
The GPT-4o mini audit uses the same information and decision fields with
temperature 0. In both audits, stylistic or abstraction changes alone do not
count as preservation failures; the central event or outcome must be changed,
contradicted, or omitted. Human annotations are the primary spot check, while
the LLM annotations provide a secondary audit of the same sampled contrasts.

\subsection{Dataset Judge Disagreement Cases}
\label{app:judge-disagreements}

This subsection explains why the LLM judge is treated as secondary audit
evidence rather than the primary dataset-quality signal. We inspected
overlapping cases where the human audit marked the assignment, event, and
sentiment as correct but GPT-4o mini disagreed on at least one field. The eight
disagreements fall into two patterns: occasional assignment-field reversals and
conservative preservation judgments. Table~\ref{tab:judge-disagreement-examples}
shows one example of each pattern.

\begin{table*}[t]
\centering
\footnotesize
\setlength{\tabcolsep}{3pt}
\begin{tabular}{@{}p{0.12\linewidth}p{0.31\linewidth}p{0.31\linewidth}p{0.20\linewidth}@{}}
\toprule
Usage & Text A & Text B & GPT-4o mini disagreement \\
\midrule
Tone &
The theater's film festival, held over the weekend, faced criticism from
attendees due to its limited selection and lack of engaging discussions. &
The theater's film festival over the weekend was a letdown. Attendees were
really disappointed by the limited selection and lack of engaging discussions. &
The judge reversed the \emph{formal} vs. \emph{enthusiastic} assignment, even
though its explanation described A as formal and B as enthusiastic. \\
\addlinespace
Audience &
The movie premiered at the local cinema last night. It was hard to follow and
didn't look that great, making it less fun. &
The movie premiered at the local cinema last night, but it suffered from
narrative incoherence and uneven visual execution, diminishing the viewer
experience. &
The judge marked sentiment preservation false for the \emph{child-friendly}
text, treating simplification as loss of the negative outcome. \\
\bottomrule
\end{tabular}
\caption{Representative cases where the human audit passed all fields but the
LLM judge disagreed. These cases motivate treating GPT-4o mini validation as
secondary audit evidence rather than the primary quality signal.}
\label{tab:judge-disagreement-examples}
\end{table*}

\section{Dataset Generation Prompts}
\label{app:generation-prompts}

This appendix records the prompt templates used to produce the controlled
dataset, so that the dataset-generation procedure is inspectable rather than
only described at a high level. The controlled data were generated with GPT-4o
using the prompt templates in Tables~\ref{tab:generation-prompts-main} and
\ref{tab:generation-prompts-additional}. Dynamic JSON blocks are shown as
placeholders; in the actual calls, these blocks list the current skeleton batch
or target rows and are passed verbatim to the model.

\begin{table*}[t]
\centering
\small
\begin{tabular}{@{}p{0.18\linewidth}p{0.78\linewidth}@{}}
\toprule
\multicolumn{2}{@{}l}{\textbf{Core generation prompt templates}} \\
\midrule
\textbf{Skeleton generation}
&
\textsc{System.} Generate controlled data for NLP representation analysis.
Output valid JSONL only; do not include markdown, numbering, explanations, or
extra text.
\newline
\textsc{User.} Generate exactly $N$ neutral event skeletons. Each object must
contain \emph{skeleton id}, \emph{domain}, \emph{base event}, \emph{positive
frame}, and \emph{negative frame}. Follow the requested id range and domain
schedule. Make the base event neutral, factual, and preferably 10--16 words.
Make the positive and negative frames semantically parallel, evaluate the same
outcome dimension, avoid one-sided entities, avoid explicit usage styles, avoid
sentiment words in the base event, prefer concrete everyday events, and output
JSONL only. \\
\midrule
\textbf{Tone realization}
&
\textsc{System.} Generate controlled natural-language examples for NLP
representation analysis. Output valid JSONL only; do not include markdown,
numbering, explanations, or extra text.
\newline
\textsc{User.} Given \emph{[skeleton batch]} and \emph{[target rows]}, generate
exactly one text field per target row. The tone values are \emph{formal},
\emph{casual}, \emph{enthusiastic}, and \emph{restrained}, defined as polished
professional precise wording, conversational relaxed wording, energetic but
factually grounded wording, and understated measured wording. Preserve all
target metadata, add only \emph{text}, express the base event and sentiment
frame, keep the event, entity, and outcome fixed, and change only the requested
communicative realization. Each text must be 15--35 words and 1--2 sentences.
Do not include markdown, numbering, placeholders, names, hashtags, emojis,
greetings, sign-offs, meta text, or explicit label words. \\
\midrule
\textbf{Audience realization}
&
\textsc{System.} Generate controlled natural-language examples for NLP
representation analysis. Output valid JSON only; do not include markdown,
numbering, explanations, or extra text.
\newline
\textsc{User.} Given \emph{[audience definitions]}, \emph{[skeleton batch]},
and \emph{[target rows]}, generate exactly one text field per target row.
The audience definitions specify background-knowledge readers for
\emph{specialist-facing}, ordinary adult readers and low-jargon wording for
\emph{general-public-facing}, short simple concrete wording for
\emph{child-friendly}, and adult peer, classmate, coworker, or familiar-reader
wording for \emph{peer-facing}.
Preserve all target metadata and add only \emph{text}. All audience versions
for the same skeleton and sentiment must express the same base event, sentiment
frame, process constraint, and impact. Versions may differ only in lexical
choice, abstraction level, terminology, explanation level, and action
orientation. Keep the event, actor, timing, polarity, and outcome fixed. Do not
add remedies, causes, decisions, approvals, budgets, names, numbers, concrete
facts, or downstream effects not present in the skeleton. Each text must be
18--35 words and 1--2 sentences. \\
\bottomrule
\end{tabular}
\caption{Core generation prompt templates. Dynamic JSON inputs are abbreviated
with bracketed placeholders for readability.}
\label{tab:generation-prompts-main}
\end{table*}

\begin{table*}[t]
\centering
\small
\begin{tabular}{@{}p{0.20\linewidth}p{0.76\linewidth}@{}}
\toprule
\multicolumn{2}{@{}l}{\textbf{Additional prompt constraints}} \\
\midrule
\textbf{Specialist-facing rule}
&
Use precise domain terms only to paraphrase
facts already present in the skeleton. Do not add new metrics, mechanisms,
risks, priorities, causes, remedies, decisions, downstream effects, or
performance dimensions; do not make the text specialist-facing by adding causal
analysis. \\
\midrule
\textbf{Peer-facing rule}
&
Write for an adult peer, classmate, coworker, or
familiar reader. Keep the same event, actor, timing, polarity, and outcome. Do
not add advice, recommendations, future consequences, risks, priorities,
remedies, decisions, new facts, technical terminology, or child-directed
wording. \\
\midrule
\textbf{Repair pass}
&
\textsc{System}: Repair controlled NLP dataset examples
and return valid JSON only. \textsc{User}: rewrite only the text fields that
failed QC while preserving the same event, sentiment, usage type, usage value,
and metadata. Each repaired text must be 15--35 words and 1--2 sentences,
express the base event and sentiment frame, preserve the requested usage style
without naming it, and avoid labels, meta words, hashtags, emojis, greetings,
sign-offs, ``Dear diary,'' ``Sure,'' and ``Here's.'' \\
\bottomrule
\end{tabular}
\caption{Additional prompt constraints for audience generation and QC repair.}
\label{tab:generation-prompts-additional}
\end{table*}

\section{Model and Representation Extraction Details}
\label{app:model-details}

This appendix fixes the model and representation-extraction choices used by the
main analyses. These details matter because all geometry, erasure, and steering
results depend on the checkpoint, layer indexing, pooling choice, and hidden
state extraction protocol. In the main text, we use the short names Llama,
Mistral, and Gemma for the three instruction-tuned decoder-only checkpoints in
Table~\ref{tab:model-checkpoints}.

\begin{center}
\begin{minipage}{\linewidth}
\centering
\scriptsize
\setlength{\tabcolsep}{2pt}
\begin{tabular}{@{}p{0.17\linewidth}p{0.42\linewidth}p{0.33\linewidth}@{}}
\toprule
Model & Checkpoint & Key details \\
\midrule
Llama &
Meta-Llama-3-8B-Instruct &
32 layers; hidden 4096; 32 heads; 8k context \\
Mistral &
Mistral-7B-Instruct-v0.2 &
32 layers; hidden 4096; 32 heads; grouped-query attention; 32k context \\
Gemma &
Gemma-7B-IT &
28 layers; hidden 3072; 16 heads; 8k context \\
\bottomrule
\end{tabular}
\captionof{table}{Model checkpoints and architectural details. References are Meta
Llama~3~\citep{grattafiori2024llama}, Mistral~7B~\citep{jiang2023mistral7b}, and
Gemma~\citep{team2024gemma}.}
\label{tab:model-checkpoints}
\end{minipage}
\end{center}

For representation extraction, we encode only the controlled text itself. We
use the tokenizer associated with each checkpoint and keep its default
special-token behavior. Padding uses the EOS token when a checkpoint lacks a
separate pad token, and sequences are padded or truncated to 128 tokens. We
extract all transformer layers with \texttt{output\_hidden\_states=True};
hidden states are computed in bfloat16 for Llama and float16 for Mistral and
Gemma, then saved as float32 arrays for analysis. Local model inference and
activation extraction use NVIDIA H100 hardware. We do not fine-tune any model:
the representation experiments use forward-pass activations from the public
checkpoints, and the generation experiments modify hidden states only at
inference time. Unless otherwise specified, data splits, shuffled baselines,
and randomized evaluation order use seed 2026. Generation-specific decoding
parameters are reported in Appendix~\ref{app:generation-diagnostic-details}.

\section{Additional Shared-Direction and Residual-Usage Results}
\label{app:residual-geometry-details}

\begin{figure*}[t]
  \centering
  \includegraphics[width=0.97\textwidth]{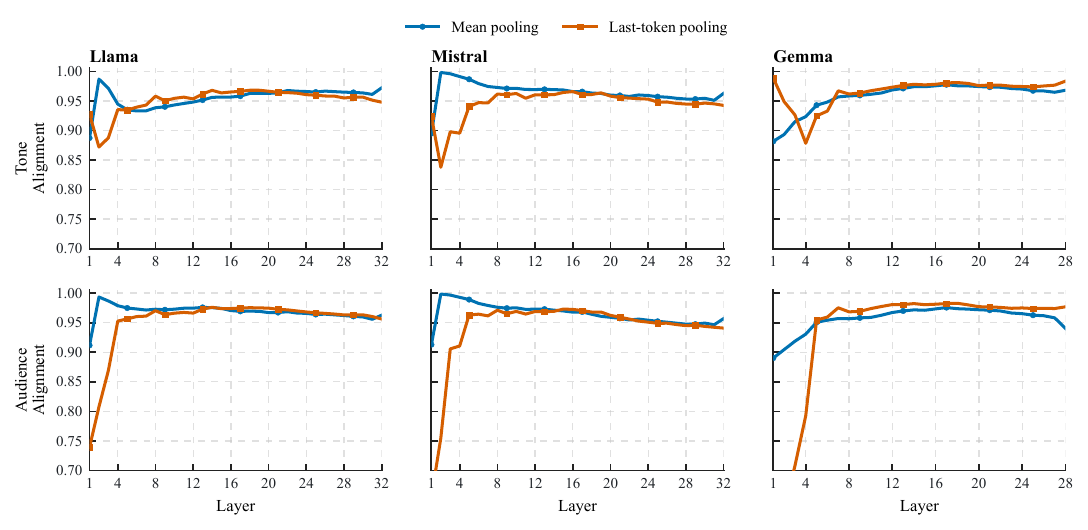}
  \caption{Layer-wise alignment with the shared sentiment direction. Each
  point reports the cosine alignment between the held-out usage-family mean
  positive--negative representation delta and the sentiment direction estimated
  from the training skeletons. Rows separate tone and audience families, columns
  separate models, and colors separate mean pooling from last-token pooling.
  High alignment across families and layers supports the shared sentiment
  direction assumption summarized in Table~\ref{tab:shared-residual-summary}.}
  \label{fig:backbone-alignment-layerwise}
\end{figure*}

\begin{figure*}[t]
  \centering
  \includegraphics[width=0.97\textwidth]{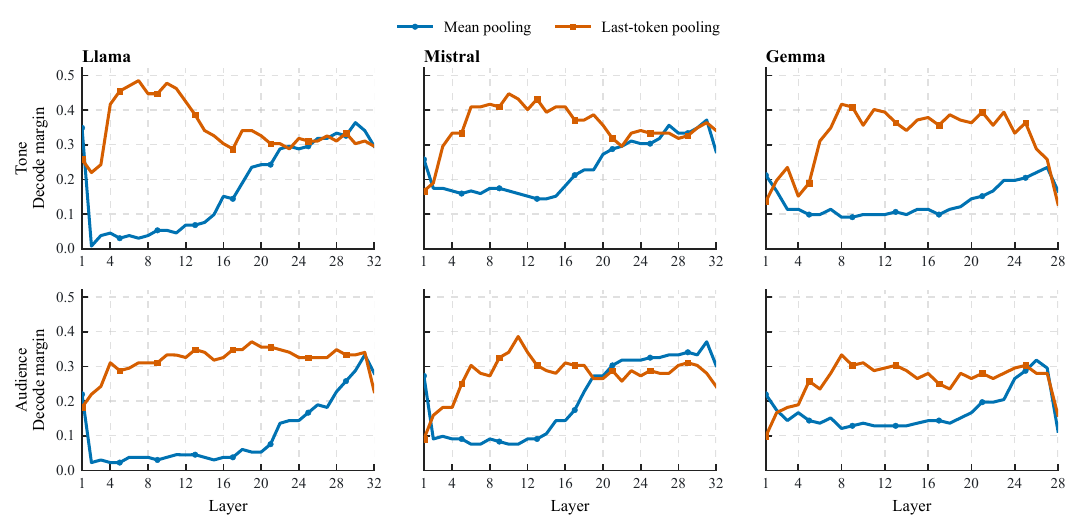}
\caption{Layer-wise residual usage decoding after removing the shared
sentiment direction. Each point reports the nearest-centroid usage-decoding
margin above the 95th percentile of the within-skeleton shuffled-label
baseline; higher values mean that held-out instances are closer to the
  centroid of their own usage value. Blue lines use mean pooling, and orange
  lines use last-token pooling. Mean pooling generally increases toward later
  layers, whereas last-token pooling is already strong in early-to-middle
layers and is flatter afterward.}
\label{fig:residual-usage-layerwise}
\end{figure*}

This appendix expands the aggregate geometry results from the main text into
layer-wise diagnostics. The goal is to show that the shared-direction and
residual-usage conclusions are not driven by a single selected layer. Figures~\ref{fig:backbone-alignment-layerwise}
and \ref{fig:residual-usage-layerwise} expand the shared-direction and
residual-usage results summarized in Table~\ref{tab:shared-residual-summary}
without aggregating across layers. Figure~\ref{fig:backbone-alignment-layerwise}
plots the close alignment between held-out tone and audience mean deltas and the
train-estimated shared sentiment direction across models and layers. Mean
pooling is weak in early layers for residual usage decoding and becomes more
separable toward later layers. Last-token pooling shows stronger usage
separation in early-to-middle layers and then remains flatter, with a sharper
late-layer drop for Gemma than for Llama or Mistral. The skeleton-level
train/development/test split uses seed 2026.

Together, these layer-wise plots provide two checks that are hidden by the
main-text medians. First, the shared sentiment direction is not a single-layer
artifact: held-out tone and audience deltas remain closely aligned with the
train-estimated direction across most layers and all three models. Second, the
residual usage signal persists after removing that direction, but its layer
profile depends on pooling. This is why the main text reports both pooling
choices for the shared/residual decomposition and uses last-token residuals for
the later steering-aligned analyses.

\section{Additional Residual-Erasure Results}
\label{app:erasure-details}

This appendix reports layer-level checks for the targeted residual-erasure
experiment. These checks support the main-text claim that the learned residual
subspaces carry held-out usage information rather than merely reducing the
overall sentiment delta. Table~\ref{tab:erasure-diagnostics-app} reports
layer-level diagnostics for the last-token erasure experiment summarized in
Table~\ref{tab:targeted-erasure-results}.

\begin{table*}[t]
\centering
\small
\setlength{\tabcolsep}{3pt}
\begin{tabular}{@{}llrrrr@{}}
\toprule
Model & Usage & Layers & Target$>$Rand & Target$>$Shuf & Target$>$Other \\
\midrule
Llama & Tone & 32 & 1.000 & 1.000 & 1.000 \\
Llama & Audience & 32 & 1.000 & 1.000 & 1.000 \\
Mistral & Tone & 32 & 1.000 & 1.000 & 1.000 \\
Mistral & Audience & 32 & 1.000 & 1.000 & 1.000 \\
Gemma & Tone & 28 & 1.000 & 1.000 & 1.000 \\
Gemma & Audience & 28 & 1.000 & 1.000 & 1.000 \\
\bottomrule
\end{tabular}
\caption{Layer-level support for last-token targeted residual erasure. Values
are fractions of layers. Target$>$Rand and Target$>$Shuf report the fraction
of layers where target erasure reduces the target usage metrics more than
random rank-3 residual subspace erasure and label-shuffled same-family rank-3
erasure, respectively.
Target$>$Other reports the fraction of layers where target erasure reduces the
target usage family more than the non-target usage family. The shared-direction
projection is retained above 0.95 in all layers for all rows.}
\label{tab:erasure-diagnostics-app}
\end{table*}

Across all evaluated layers, target erasure reduces the corresponding usage
metrics more than random, shuffled, and other-family controls while preserving
the shared sentiment projection.

\section{Generation-Time Steering Details and Robustness Checks}
\label{app:generation-diagnostic-details}

This appendix documents the evaluation protocol behind the judge-based
generation results, so that the main-text win/tie/loss counts can be traced to
blind pairwise comparisons and separate sanity checks.
Table~\ref{tab:generation-diagnostic-breakdown} gives the per-control breakdown
for the Llama generation-time steering experiment in
Section~\ref{sec:generation-diagnostic}. The setting is fixed at layer 20 with
$\beta=50$. Generation uses seed 2026, bfloat16 model weights, a batch size of
8, at most 48 new tokens, temperature 0.7, and top-$p$ 0.9. The
source-conditioned prompt names the source tone, event, and
outcome frame, but not the target tone. The judge prompt specifies the target
tone for evaluation while hiding the intervention condition and presentation
order.

For pairwise tone evaluation, the judge sees the requested situation, target
tone, source tone to reduce, requested sentiment, and two blind-labeled outputs.
The intervention labels are hidden, and A/B order is randomized with seed 2026
when judge pairs are created. The judge then decides which output better matches
the target tone while also checking source-tone reduction, preservation of the
requested sentiment and event content, fluency, and artifacts or instruction
leakage. We run two additional binary checks outside this pairwise comparison:
the source gate evaluates no-intervention outputs, and the polarity check
evaluates target-injected outputs.

The per-control breakdown clarifies what the pooled win/tie/loss counts in the
main text are measuring. Target injection beats the no-intervention and random
controls almost uniformly, which rules out the explanation that any residual
direction with the same scale would produce the effect. The wrong-target and
opposite-target controls are more stringent: they test whether the injected
direction points toward the requested target tone rather than merely making the
output more edited or stylistically different. The restrained target is the
hardest condition under the wrong-target control, but it still preserves the
overall pattern of target-specific steering.

The examples show the qualitative behavior behind the judge counts: the event
and negative outcome remain fixed, while wording moves along the requested tone
contrast. The quantitative evaluation in
Table~\ref{tab:generation-diagnostic-breakdown} provides the main evidence for
the steering result.

\subsection{Independent-Judge Re-evaluation}

We submit the same blind pairs to an independent Claude Sonnet 5 judge, without
regenerating or filtering outputs. Table~\ref{tab:claude-tone-reeval} reports
the resulting target-tone preferences. The target direction also remains
preferred when results are grouped by control: the W/T/L counts are 141/11/0
against no intervention, 143/9/0 against a random residual direction, 138/8/6
against a wrong target, and 147/5/0 against the opposite target. Thus, the
independent judge changes individual decisions but preserves the overall
conclusion.

\begin{table}[H]
\centering
\small
\setlength{\tabcolsep}{3pt}
\begin{tabular}{@{}lrrr@{}}
\toprule
Source $\rightarrow$ target & $N$ & W/T/L & Win \% \\
\midrule
Pooled & 608 & 569/33/6 & 93.6 \\
formal $\rightarrow$ casual & 160 & 157/1/2 & 98.1 \\
restr. $\rightarrow$ enth. & 160 & 130/30/0 & 81.2 \\
casual $\rightarrow$ formal & 160 & 160/0/0 & 100.0 \\
enth. $\rightarrow$ restr. & 128 & 122/2/4 & 95.3 \\
\bottomrule
\end{tabular}
\caption{Independent Claude Sonnet 5 re-evaluation of the fixed Llama tone-steering
outputs. W/T/L denotes target-injection wins, ties, and losses. Win percentages
retain ties in the denominator.}
\label{tab:claude-tone-reeval}
\end{table}

\subsection{Human-Written Transfer}

One human annotator wrote 40 sentences from 10 held-out event skeletons, with one
formal and one casual version for each positive and negative outcome. No
human-written sentence is used to estimate the direction. We compare the signed projection of each
within-skeleton, within-sentiment formal--casual pair onto the layer-20
direction estimated from the controlled training data. Table~\ref{tab:rewrite-transfer}
reports both the tested direction and a usage-projected control. For the tested
direction, positive and negative items each contribute 9 correct orderings out
of 10. The empirical $p$-value is the fraction of 2,000 matched random
directions with at least as many wins.

\begin{table}[H]
\centering
\small
\setlength{\tabcolsep}{3pt}
\begin{tabular}{@{}lrrr@{}}
\toprule
Direction & Correct & Accuracy & Random $p$ \\
\midrule
Layer-20 tone direction & 18/20 & 90.0 & .0045 \\
Usage-projected control & 9/20 & 45.0 & .6735 \\
\bottomrule
\end{tabular}
\caption{Zero-shot formal--casual ordering on the controlled human-written
set. Directions are estimated without using these examples.}
\label{tab:rewrite-transfer}
\end{table}

\subsection{Audience Steering}

For the audience evaluation, we use Llama layer 24 with $\beta=40$, greedy
decoding, batch size 1, and 20 held-out specialist-facing prompts. The prompt
contains the event and outcome facts but not the source or target audience
label. A blind Claude Sonnet 5 judge evaluates audience style; outcome preservation and
surface-form artifacts are audited separately. Table~\ref{tab:audience-steering}
shows that direct child-friendly injection is preferred to both a matched
random direction and no intervention. Artifacts appear in both target and
random-direction outputs, while the target direction achieves higher
child-friendly preference than both controls and preserves the requested
outcome.

\begin{table}[H]
\centering
\small
\setlength{\tabcolsep}{2.5pt}
\begin{tabular}{@{}lrrr@{}}
\toprule
Comparison & Child pref. & Pres. loss & Art. \\
\midrule
Target vs. random & 19/20 & 0/20 & 5/20 \\
Target vs. none & 18/20 & 0/20 & 5/20 \\
Random vs. none & 12/20 & 1/20 & 4/20 \\
\bottomrule
\end{tabular}
\caption{Held-out specialist-facing $\rightarrow$ child-friendly audience
steering results. Artifacts are counted on the first condition in each comparison.}
\label{tab:audience-steering}
\end{table}

\begin{table*}[t]
\centering
\small
\setlength{\tabcolsep}{3pt}
\begin{tabular}{@{}lllcc@{}}
\toprule
Source $\rightarrow$ target & Control & Target W/T/L & Polarity & Art. \\
\midrule
formal $\rightarrow$ casual & none & 40/0/0 & 40/40 & 1 \\
formal $\rightarrow$ casual & random & 40/0/0 & 40/40 & 1 \\
formal $\rightarrow$ casual & wrong target & 38/0/2 & 40/40 & 1 \\
formal $\rightarrow$ casual & opposite & 40/0/0 & 40/40 & 1 \\
\addlinespace
restrained $\rightarrow$ enthusiastic & none & 30/9/1 & 39/40 & 0 \\
restrained $\rightarrow$ enthusiastic & random & 30/9/1 & 39/40 & 0 \\
restrained $\rightarrow$ enthusiastic & wrong target & 34/5/1 & 39/40 & 0 \\
restrained $\rightarrow$ enthusiastic & opposite & 34/6/0 & 39/40 & 0 \\
\addlinespace
casual $\rightarrow$ formal & none & 40/0/0 & 40/40 & 0 \\
casual $\rightarrow$ formal & random & 40/0/0 & 40/40 & 0 \\
casual $\rightarrow$ formal & wrong target & 40/0/0 & 40/40 & 0 \\
casual $\rightarrow$ formal & opposite & 40/0/0 & 40/40 & 0 \\
\addlinespace
enthusiastic $\rightarrow$ restrained & none & 32/0/0 & 31/32 & 0 \\
enthusiastic $\rightarrow$ restrained & random & 32/0/0 & 31/32 & 0 \\
enthusiastic $\rightarrow$ restrained & wrong target & 22/1/9 & 31/32 & 0 \\
enthusiastic $\rightarrow$ restrained & opposite & 32/0/0 & 31/32 & 0 \\
\bottomrule
\end{tabular}
\caption{Per-control breakdown for source-conditioned generation-time tone
steering. Target W/T/L compares target injection with one control condition.
Polarity and Art. are repeated within each source--target pair because they are
computed on target-injected outputs rather than pairwise comparisons.}
\label{tab:generation-diagnostic-breakdown}
\end{table*}

\begin{table*}[t]
\centering
\footnotesize
\setlength{\tabcolsep}{3pt}
\begin{tabular}{@{}p{0.18\linewidth}p{0.38\linewidth}p{0.38\linewidth}@{}}
\toprule
Source $\rightarrow$ target & No intervention & Target injection \\
\midrule
formal $\rightarrow$ casual &
It is regrettable that, despite being submitted on time, the team's project
report did not meet the expected standards, as it lacked organization and
failed to incorporate key requirements. &
The team submitted the report on Friday, but unfortunately it's been noticed
that it's a bit disorganized and doesn't cover all the requirements we were
expecting. \\
\addlinespace
casual $\rightarrow$ formal &
So, we finally got the project report in on Friday afternoon, but unfortunately
it's a bit of a mess, disorganized and missing some key stuff. &
The team's submission of the project report, which was completed on Friday
afternoon, unfortunately, fell short due to poor organization and failure to
address key requirements. \\
\bottomrule
\end{tabular}
\caption{Example Llama generations under tone steering. Within each row, the
event and negative outcome frame are held fixed. The no-intervention output
follows the prompted source tone, while target injection shifts wording toward
the target tone while preserving the negative outcome.}
\label{tab:generation-examples}
\end{table*}

\end{document}